\documentclass[10pt,conference,a4paper]{IEEEtran}
\IEEEoverridecommandlockouts
\usepackage{cite}
\usepackage{amsmath,amssymb,amsfonts}
\usepackage{algorithmic}
\usepackage{graphicx}
\usepackage{textcomp}
\usepackage{xcolor}
\usepackage{multirow}
\def\BibTeX{{\rm B\kern-.05em{\sc i\kern-.025em b}\kern-.08em
    T\kern-.1667em\lower.7ex\hbox{E}\kern-.125emX}}
\begin{document}

\title{Feathers dataset for Fine-Grained Visual Categorization\\
}

\author{\IEEEauthorblockN{Alina Belko}
\IEEEauthorblockA{
\textit{Samara University}\\
Samara, Russia \\
alinabelko@gmail.com}
\and
\IEEEauthorblockN{Konstantin Dobratulin}
\IEEEauthorblockA{
\textit{Samara University}\\
Samara, Russia \\
zsxoff@gmail.com}
\and
\IEEEauthorblockN{Andrey Kuznetsov}
\IEEEauthorblockA{
\textit{Samara University}\\
Samara, Russia \\
kuznetsoff.andrey@gmail.com}
}

\maketitle

\begin{abstract}
This paper introduces a novel dataset FeatherV1, containing 28,272 images of feathers categorized by 595 bird species. It was created to perform taxonomic identification of bird species by a single feather, which can be applied in amateur and professional ornithology. FeatherV1 is the first publicly available bird's plumage dataset for machine learning, and it can raise interest for a new task in fine-grained visual recognition domain. The latest version of the dataset can be downloaded at https://github.com/feathers-dataset/feathersv1-dataset. We also present feathers classification task results. We selected several deep learning architectures (DenseNet based) for categorical crossentropy values comparison on the provided dataset.
\end{abstract}


\section{Introduction}
\begin{figure*}[t!]
\centerline{\includegraphics[width=\textwidth]{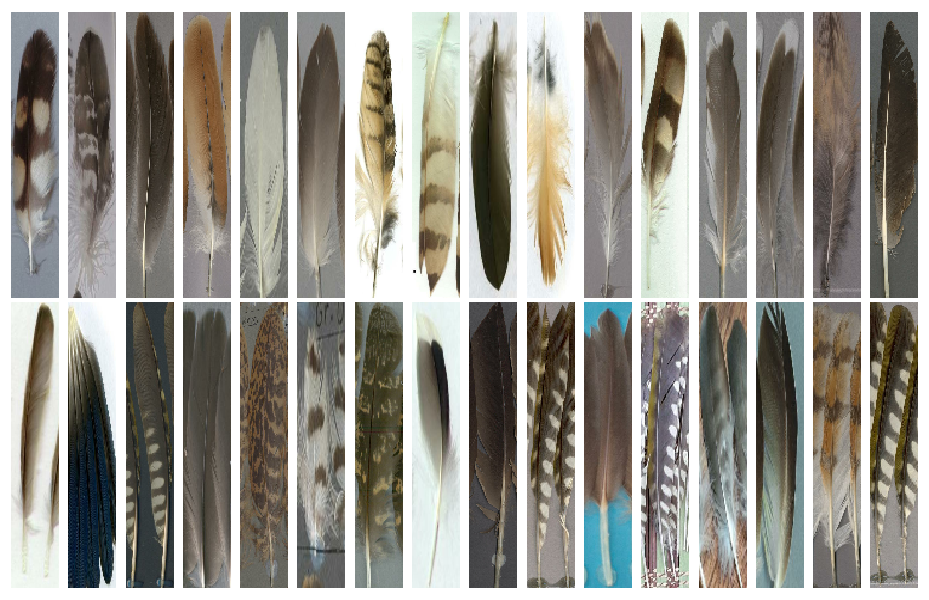}}
\caption{FeathersV1 sample images after normalization.}
\label{dataset_example}
\end{figure*}

Feathers remained out of computer vision focus, despite a sufficient amount of unstructured data available on the Internet. The lack of high-quality datasets may limit the research progress of the community in this domain. Bird
species categorization can be called a classic fine-grained visual recognition task\cite{Caltech-UCSD,Birdsnap,Deep_Nets,Look_closer}. However, as far as we know, there were no previous attempts to categorize bird species by an image of a single feather.
In this paper, we introduce our on-going dataset project called FeathersV1, aimed at studying the problem of fine-grained classification and recognition of bird species by a feather. The dataset contains 28,272 feather images of 595 bird species. Several enthusiasts provided initial images, a full list of authors can be found at the GitHub page of the dataset. We manually divided every image into multiple images with a single feather and constructed the result dataset. Several examples of the dataset images are provided in Fig.~\ref{dataset_example}.

\begin{figure}[b!]
\centerline{\includegraphics[width=\columnwidth]{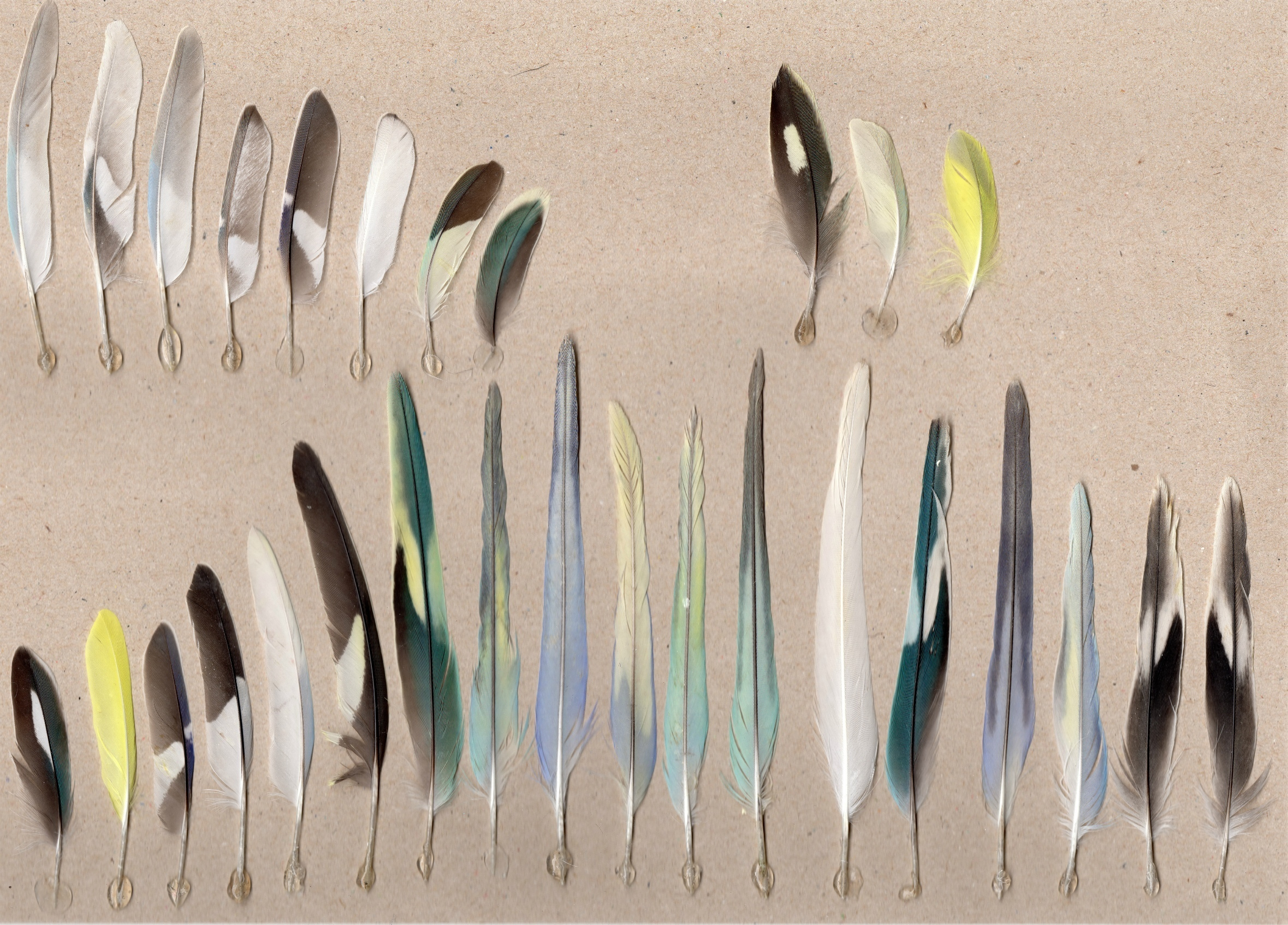}}
\caption{An example of color and shape variation of Melopsittacus undulatus.}
\label{budgie}
\end{figure}

Feathers are not standard objects in fine-grained visual classification problems, although they have some interesting aspects for this task. Every feather has the same overall structure: calamus, rachis, barbs, and afterfeather, but they can greatly vary both in intraspecific and interspecific ways. 
The dataset can be called challenging for multiple reasons. Firstly, all of the feathers have a relatively similar structure, and the difference between feathers of different species can be low, especially for species of the same order and even genus. Secondly, their intraspecific differences can be relatively high due to biological reasons. Many bird species have sexual dimorphism, age-related differences, and morphological deviations, such as melanism or albinism \cite{Deviations}. Some species have multiple color variations, which can be caused by natural reasons or by breeders, example of various feathers of budgerigars is presented in Fig.~\ref{budgie}. Even feathers of a single specimen can vary significantly in shape and color depending on type: wing (remiges), tail (rectrices), contour (coverts), or semiplume \cite{Feather structure}. 

Among ornithology tasks where the created dataset can be applied is aviation ornithology. One of the main steps for risk management of birdstrikes is to identify a species of bird that caused an incident. After collisions between birds and aircraft, small feather fragments are usually the only evidence that a bird was involved in the incident. One of the essential steps in reducing the hazard of bird strikes is appropriately identifying which species cause incidents and pose a higher risk to aviation. Major airports use molecular means of identifying species from a variety of tissue types, including blood, muscle, skin, and feathers. Following collisions between birds and aircraft, several techniques are available to identify birds involved in aircraft collisions \cite{BRIS}. Major airports use expertise, and laboratory assistance, which includes microscopic examination of downy barbules of feathers, electrophoresis identification, and DNA examination \cite{DNA, Microscopic}. However, in many cases, unaided visual identification is still in use because not all airports have required laboratory equipment. 

FeathersV1 allows performing identification without specialized knowledge and laboratory equipment so that it can be used in the field. Other possible applications are amateur ornithology, bird watching, and feather collecting. The ability to identify feathers without requesting help from other people can raise interest in amateur ornithology because of people's curiosity to accidentally found feathers.

Our contributions are three-fold. First, we introduce a new dataset of feather images with species annotations. Second, we describe how data was collected using online-resources and the work of hobbyists and enthusiasts. Third, we present baseline classification results on species identification using several DenseNet convolutional neural networks (CNN) architectures. Sect. \ref{Dataset} describes the content of the FeathersV1 dataset, and it is properties. The following Sect. \ref{Dataset construction} describes the dataset construction and preprocessing. In Sect. \ref{Baseline} we examine the performance of a baseline classifier on the data. Finally, Sect. \ref{Summary} contains a results summary and future research plans. 

\section{Dataset} \label{Dataset}
FeathersV1 contains 28,272 images of feathers annotated with their species. Images are organized in a two-level hierarchy:
\begin{itemize}
\item \textbf{Species.} This is the most specific class label with Latin names of bird species. The dataset contains 595 species of birds.
\item \textbf{Order.} This level represents the biological order of bird species. It is not used in the classification task but provides a more organized structure. The dataset contains 23 orders of birds.
\end{itemize}

\begin{figure}[b!]
\centerline{\includegraphics[width=\columnwidth]{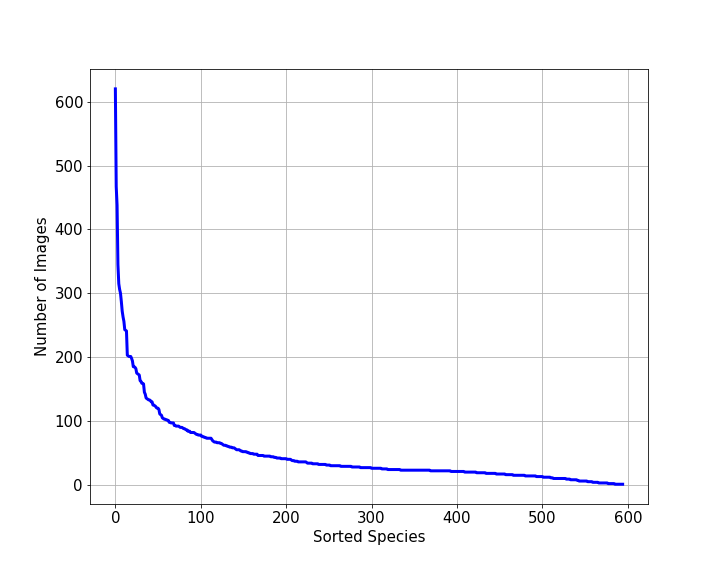}}
\caption{Quantitative distribution of images per species.}
\label{distribution}
\end{figure}

One of the main challenges of this dataset is its very uneven diversification of images per class. It may vary between 2 and 620, and depends on species' abundance and popularity among collectors. Due to this aspect, it requires a high augmentation level on some of the classes to exclude their imbalance. To make the dataset more even, we split the dataset into three datasets, containing Top-50, Top-100, and all classes ordered by images number. Quantitative distribution of images per species is shown in Fig.~\ref{distribution}.

\begin{table}
\caption{Quantity distribution of images in Orders}
\centering
\begin{tabular}{l|l|l|l}
Orders              & Images & Species & Num. images per
species  \\ 
\hline
Passeriformes       & 8~480  & 252     & 33,7                     \\
Anseriformes        & 4~555  & 62      & 73,5                     \\
Charadriiformes     & 3~908  & 45      & 86,8                     \\
Accipitriformes     & 2~783  & 41      & 67,9                     \\
Strigiformes        & 1~650  & 24      & 68,8                     \\
Galliformes         & 1~104  & 28      & 39,4                     \\
Piciformes          & 872    & 22      & 39,6                     \\
Ciconiiformes       & 800    & 18      & 44,4                     \\
Caprimulgiformes    & 658    & 8       & 82,3                     \\
Columbiformes       & 656    & 19      & 34,5                     \\
Gruiformes          & 557    & 14      & 39,8                     \\
Coraciiformes       & 415    & 11      & 37,7                     \\
Apodiformes         & 411    & 9       & 45,7                     \\
Procellariiformes   & 272    & 7       & 38,9                     \\
Phoenicopteriformes & 259    & 2       & 129,5                    \\
Gaviiformes         & 239    & 4       & 59,8                     \\
Psittaciformes      & 234    & 10      & 23,4                     \\
Bucerotiformes      & 161    & 6       & 26,8                     \\
Cuculiformes        & 128    & 4       & 32,0                     \\
Pelecaniformes      & 73     & 7       & 10,4                     \\
Trogoniformes       & 26     & 1       & 26,0                     \\
Pteroclidiformes    & 19     & 1       & 19,0                     \\
Coliiformes         & 12     & 1       & 12,0                    
\end{tabular}
\label{orders_table}
\end{table}

Image quality also varies because of the different nature of images. They are collected from several enthusiasts and were made at different times. 

Dataset is divided into train and test subsets. We define classification tasks - bird species recognition. The performance is evaluated as class-normalized average classification accuracy, obtained as the average of the diagonal elements of the normalized confusion matrix \cite{Confusion}.

\begin{table}[bp!]
\caption{Test and validation split}
\begin{center}
\begin{tabular}{|c|c|c|}
\hline
\textbf{}&\multicolumn{2}{|c|}{\textbf{Count of images in categories}} \\
\cline{2-3} 
\textbf{Category} & \textbf{\textit{Train split}}& \textbf{\textit{Validation split}} \\
\hline
    Top-50 classes & 8,251 & 2,063 \\ \hline
    Top-100 classes & 11,953 & 2,988 \\ \hline
    All classes & 22,618 & 5,654 \\ \hline
\end{tabular}
\label{tab1}
\end{center}
\end{table}

The dataset is published for non-commercial purposes only at https://github.com/feathers-dataset/feathersv1-dataset. Please note that those images were provided by multiple people, and we do not hold the rights on initial images.

\section{Dataset construction} \label{Dataset construction}

Identifying the species of a bird from a
feather image is challenging for anyone, but ornithology experts, and collecting 28,272 such annotations is daunting in general. Sect.\ref{Initial data} explains how feather data was collected by feather enthusiasts. \ref{Images preprocessing} explains how data was preprocessed.

\subsection{Initial data} \label{Initial data}

\begin{figure}[b!]
\centerline{\includegraphics[width=\columnwidth]{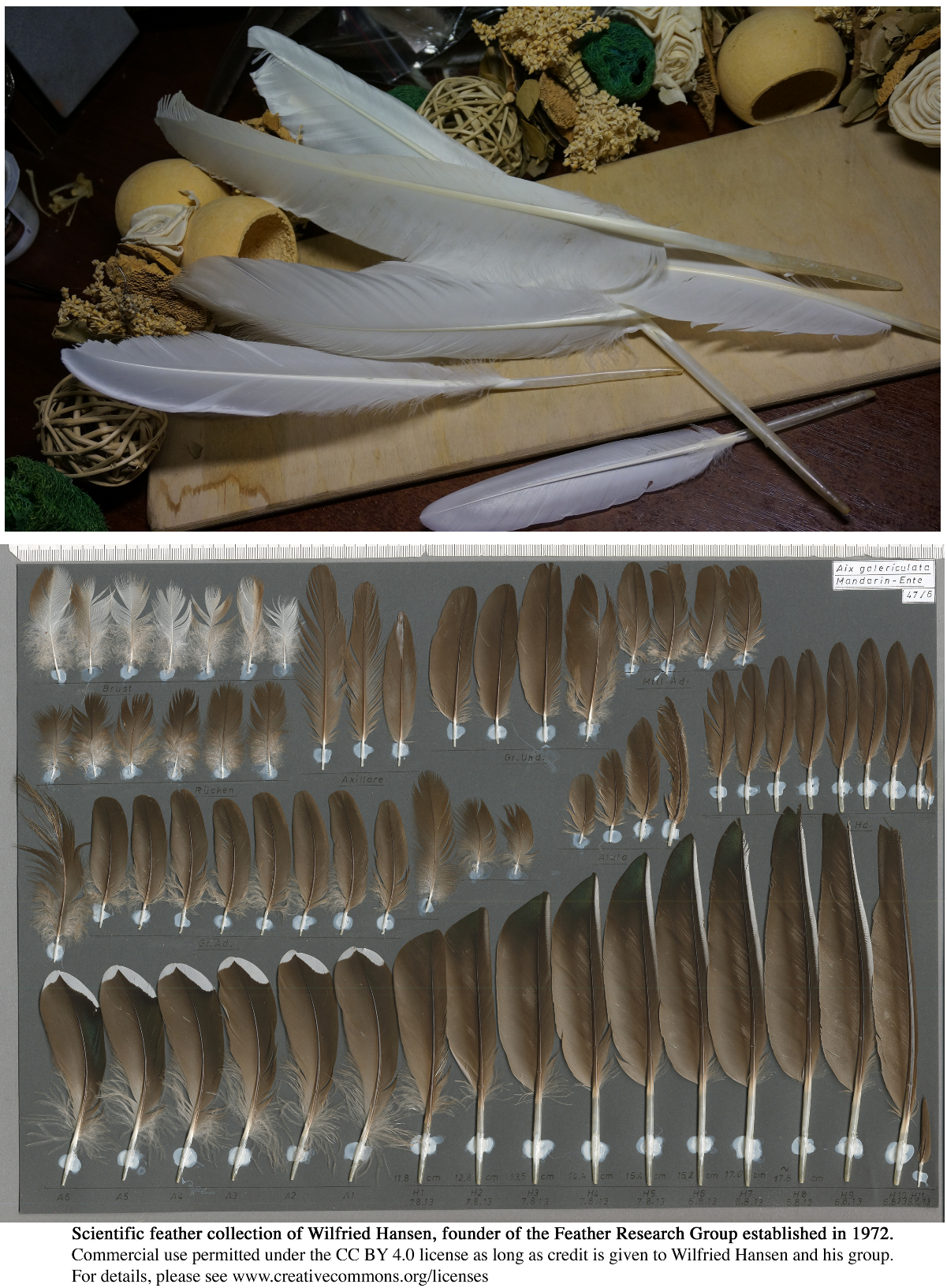}}
\caption{Example of initial images.}
\label{images_example}
\end{figure}

Feather images can be found at various sites across the Internet, where collectors publish their feather collections. Although research purposes can be considered as a fair use of images, nevertheless, we contacted collectors to ask permission to use images. We contacted several collectors, and five of them gave us permission. Also, three sites had open licenses: Creative Commons BY 4.0, Copyleft, and GNU Free Documentation License. 
To maximize a variety of images, we added images from minor collectors published on social networks. Most of these images are made for sale, and they may have some text or other objects. Example of initial images is presented in Fig.~\ref{images_example}. This allowed to add images with different lighting, angles, and background, although minor collectors usually have less variety of species and have a higher percent of incorrect recognition of feathers. 
The initial dataset contained a total of 2561 images, and the vast majority of images had many feathers, the average number of feathers per image is 11.03. 

\subsection{Images preprocessing} \label{Images preprocessing}

\begin{figure*}[t!]
\centerline{\includegraphics[width=\textwidth]{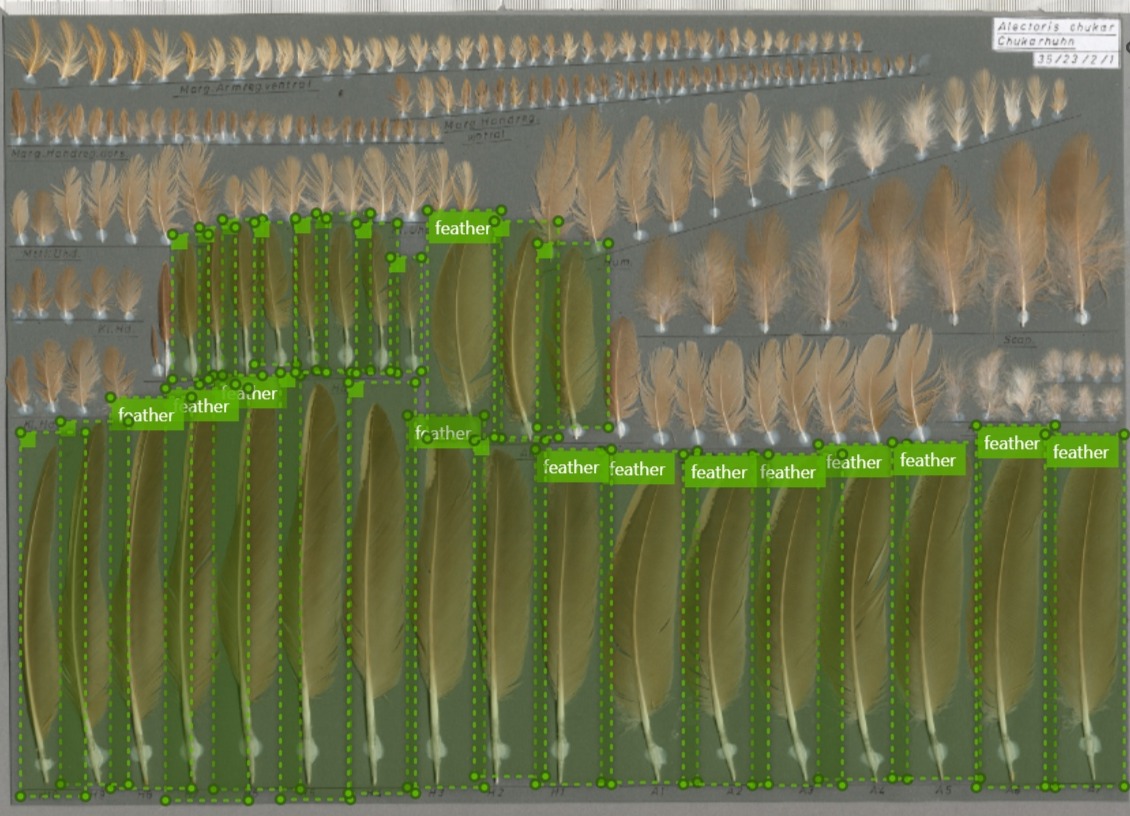}}
\caption{Initial images annotations.}
\label{bounding boxes}
\end{figure*}

After collecting images, we manually cropped every image into individual feather images to prevent the classification model from training based on feathers arrangement instead of visual patterns of individual feathers. All the images were annotated with bounding boxes via the VOTT Annotating tool. After converting annotations to JSON format, we wrote a script to save bounding boxes into individual images. Each image contains a cropped image of a single feather, sometimes with part of adjacent feathers of the same species. Dataset contains 28,272 images of 595 bird species. Not all the feathers from initial images were annotated, we filtered out down, semiplume and contour feathers if it considered not meaningful for classification, an example of bounding box annotations is presented in Fig. \ref{bounding boxes}. Also, we filtered out some of the highly overlapping feathers if they were unrecognizable for classification. 

\section{Baseline} \label{Baseline}

\begin{figure*}[h!]
\centerline{\includegraphics[width=\textwidth]{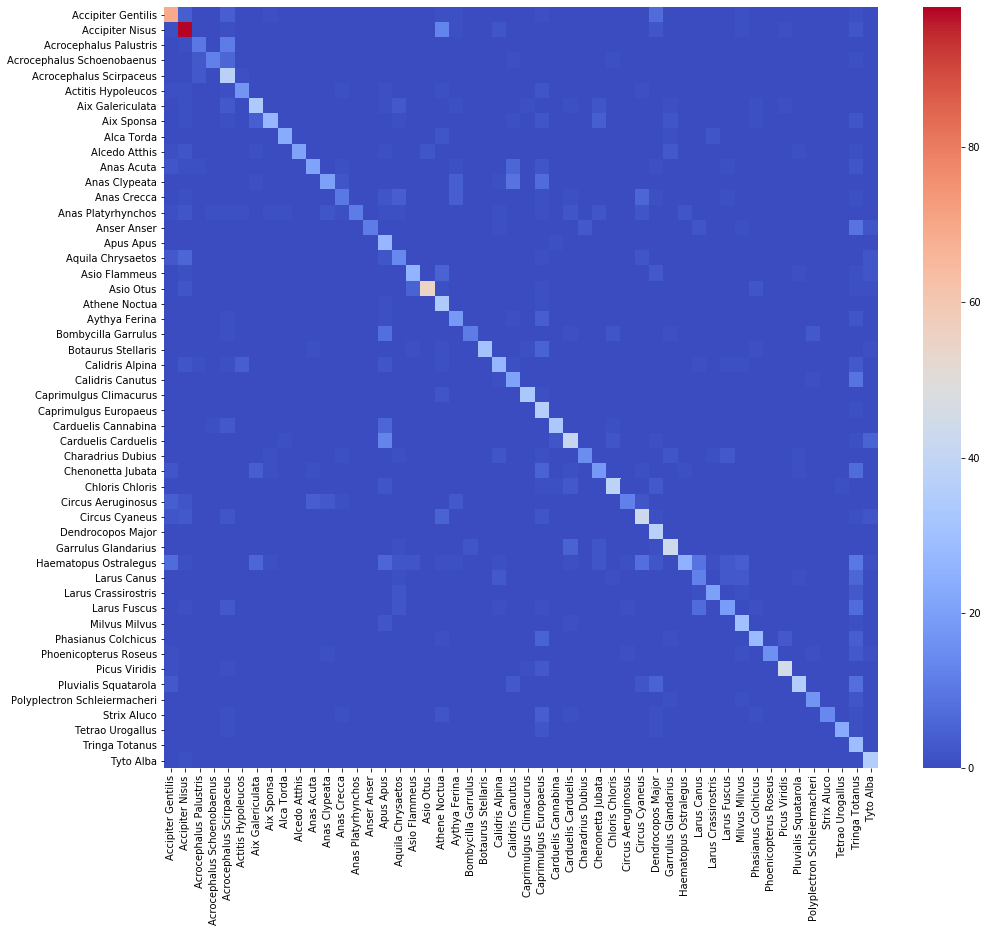}}
\caption{Confusion matrix of DenseNet169 model at Top-50 classes.}
\label{top50_confusion_matrix}
\end{figure*}

In this section, we study fine-grained feather classification using DenseNet models. We select 78,126 images
from the FeathersV1 dataset and divide them into three
subsets. The first subset (Top-50) contains
50 classes with a total of 10,314 images of bird species ordered by the number of images per class. The
second subset (Top-100) consists of 100 classes with 14,941
images in total. The last subset (All) contains the entire dataset of 595 species with 28,272 images. We compare the recognition performance of three DenseNet models: DenseNet121, DenseNet169 and DenseNet201 \cite{DenseNet}. DenseNet models show great results for fine-grained classification tasks \cite{Evaluation}. The performances of these
nine models are summarized in Table \ref{results}. Models trained on the entire dataset show the best performance, although the dataset is very imbalanced. Including into dataset many rare species with a small number of images has led to better performance, although it made the dataset more imbalanced. 
Fig. \ref{top50_confusion_matrix} shows the confusion matrix for DenseNet169 trained at Top-50 classes. This model is not one with the best performance but allows us to see results better. The model is less confident at genera of Acrocephalus and Larus because feathers of different species of those genera vary very slightly, and it is hard to recognize species manually. Recognition of Accipiter Nisus and Accipiter Gentilis is significantly more accurate because those species have 619 and 437 images, respectively, which is significantly more than other species in the dataset.
The full code of our research can be found at https://github.com/feathers-dataset/feathersv1-classification.

\begin{table*}[ht!]
\caption{Fine-grained classification results.}
\centering
\begin{tabular}{|c|c|c|c|c|} 
\hline
Subset                   & Model       & Sparse Categorical Crossentropy & Sparse Top 1 Categorical Accuracy & Sparse Top 5 Categorical Accuracy  \\ 
\hline
\multirow{3}{*}{Top-50}  & DenseNet121 & 1,4597                          & 0,6394                            & 0,8871                             \\
                         & DenseNet169 & 1,3592                          & 0,6684                            & 0,9186                             \\
                         & DenseNet201 & 1,9363                          & 0,5700                            & 0,8740                             \\ 
\hline
\multirow{3}{*}{Top-100} & DenseNet121 & 0,6771                          & 0,7989                            & 0,9709                             \\
                         & DenseNet169 & 0,7131                          & 0,7979                            & 0,9695                             \\
                         & DenseNet201 & 1,0256                          & 0,7266                            & 0,9491                             \\ 
\hline
\multirow{3}{*}{All}     & DenseNet121 & 0,8549                          & 0,7642                            & 0,9482                             \\
                         & DenseNet169 & 1,0586                          & 0,7181                            & 0,9360                             \\
                         & DenseNet201 & 0,7689                          & 0,7978                            & 0,9586                             \\
\hline
\end{tabular}
\label{results}
\end{table*}

\section{Summary} \label{Summary}
We have introduced the FeathersV1 dataset, a new dataset for fine-grained visual categorization. The data contains 28,272 feather images of 595 bird species. We believe that FeathersV1 has the potential of introducing a new kind of object to a fine-grained visual recognition domain. Feathers have interesting aspects for visual recognition.
With further development, the dataset can be used in real-world applications for professional and amateur ornithologists. In the future, we plan to increase the size of the FeathersV1 dataset by organizing image crowd-sourcing among collectors.

\section*{Acknowledgment}
Many thanks to collectors and enthusiasts who kindly provided permission to use their images or published them with open licenses. These are, Wilfried Hansen with Scientific feather collection of Wilfried Hansen, Daria Korepova from featherlab.ru, Emel'yan Kuranov, Yuliya Lygina and Yan Voron from vk.com, Michel Klemann from michelklemann.nl, Stephan Schubert from vogelfedern.de, Phil and Agi from federn.org and Christine Wedler, who kindly gave us permission to use image collection of her recently deceased husband -- Prof. Hans Schick at ornithos.de. Please note that images are
made available exclusively for non-commercial use. The original authors retain the copyright on the initial pictures and should be contacted for any
other usage of them. All the links to authors of original photos are listed at the GitHub page of the dataset.

\vspace{12pt}

\end{document}